\pdfoutput=1

\documentclass[10pt, a4paper]{article}
\usepackage[T1]{fontenc}
\usepackage{times}

\usepackage{lrec2022}
\usepackage{multibib}
\usepackage{graphicx}
\usepackage{tabularx}
\usepackage{soul}
\usepackage{adjustbox}
\usepackage{multirow}

\usepackage[utf8]{inputenc}

\usepackage{microtype}

\usepackage{hyperref}
\usepackage{comment}
\usepackage{tabularx}
\usepackage{graphicx}
\usepackage[warn]{textcomp}
\usepackage{tabularx}
\usepackage{adjustbox}
\usepackage{amssymb}
\usepackage{amsmath}
\usepackage{makecell}

\usepackage[inline]{enumitem}

\usepackage{stfloats}

\title{Cross-lingual Transfer of Monolingual Models}

\name{Evangelia Gogoulou$^\star$, Ariel Ekgren$^\circ$, Tim Isbister$^\circ$, Magnus Sahlgren$^\circ$}

\address{$^\star$RISE Research Institutes of Sweden, $^\circ$AI Sweden \\
         evangelia.gogoulou@ri.se\\
         \{ariel.ekgren, tim.isbister, magnus.sahlgren\}@ai.se\\}

\abstract{
Recent studies in cross-lingual learning using multilingual models have cast doubt on the previous hypothesis that shared vocabulary and joint pre-training are the keys to cross-lingual generalization. We introduce a method for transferring monolingual models to other languages through continuous pre-training and study the effects of such transfer from four different languages to English. Our experimental results on GLUE show that the transferred models outperform an English model trained from scratch, independently of the source language. After probing the model representations, we find that model knowledge from the source language enhances the learning of syntactic and semantic knowledge in English.
}

\begin{document}

\maketitleabstract


\section{Introduction}

Training a language model from scratch requires considerable resources, both with respect to data and computational resources. These requirements can be a limiting factor for many actors, and for smaller languages, it is not clear whether there even {\em exists} enough data to train a language model. Consequently, there is an increasing interest in using cross-lingual transfer to alleviate the requirements of training a language model from scratch. Most of the work in this direction (see the next section) concerns multilingual models.

In this paper, we want to explore the effects of cross-lingual transfer of a monolingual model into a target language space. More specifically, we want to investigate if it is possible to adapt existing monolingual models to the target language and study their downstream performance in comparison with a model trained from scratch in the target language. Additionally, we want to study the impact of language similarity between source and target language on fine-tuning performance after transfer. Based on recent work that shows that transformer-based language models encode universal properties \cite{lu2021pretrained}, we hypothesize that model knowledge learned in the source language enhances the learning of the target language independently of language proximity.

Our contributions in this work are the following: \begin{enumerate*}[label=(\textbf{\roman*)}]\item we introduce an adaptation method for cross-lingual transfer (Section \ref{sec:method}), 
\item we show that the models that have been transferred {\em to} English outperform an English model trained from scratch in the GLUE benchmark for {\em all} source languages studied here (Section \ref{subsec:glue}), 
\item through probing the model representations, we demonstrate that abstractions learned in the source language are transferred to English
(Section \ref{subsec:probing}).
\end{enumerate*}

\section{Related Work}

The standard methodology of transferring knowledge across languages is by training either cross-lingual  \cite{ruder2019survey} or multilingual models \cite{schwenk-douze-2017-learning,devlin-etal-2019-BERT,conneau-etal-2020-unsupervised,xue2020mt5,chung2021rethinking}. These latter models are trained on massively multilingual data using a shared vocabulary, which has proven to be successful for \emph{zero-shot} cross-lingual transfer \cite{pires-etal-2019-multilingual}, where the multilingual model, in this case mBERT, is fine-tuned on a downstream task in the source language and evaluated on the same task in the target language.

\newcite{pires-etal-2019-multilingual} hypothesize that zero-shot cross-lingual generalization is facilitated by using a shared vocabulary. Several recent studies contradict this assumption.
\newcite{K2020Cross-Lingual} show that in a joint-training setting, multilinguality can be achieved even if the two languages do not share any vocabulary.
\newcite{conneau-etal-2020-emerging} train jointly bilingual masked language models that share only the top two Transformer layers.
A different perspective is provided by \newcite{artetxe-etal-2020-cross}, who disregard even the joint pre-training constraint and transfer monolingual BERT to a new language by learning only a new embedding matrix from scratch while freezing the rest of the model. Their results indicate that neither shared vocabulary nor joint pre-training are necessary for cross-lingual transfer in the zero-shot setting. This method has also been applied to GPT-2 \cite{de-vries-nissim-2021-good}. 

The overarching conclusion mainly from \newcite{conneau-etal-2020-emerging} and \newcite{artetxe-etal-2020-cross} is that zero-shot cross-lingual transfer is facilitated by shared statistical properties between language spaces, rather than multilingual pre-training. We would like to explore if this holds in a different transfer scenario, where a model is transferred to a new language and fine-tuned on monolingual tasks in that language. Our hypothesis is that the statistics of language acquired by a model in the source language will transfer and boost monolingual task performance in the target language.

 \begin{table*}[!b]
 \centering
        \begin{tabular}{|l|l|l|r|r|}
        \hline
         \textbf{Language} & \textbf{Model name} & \textbf{Alias} & \textbf{Vocab size} & \textbf{Data (GB)} \\
         \hline
         English   & BERT-base (ours)       & en        & 32,000   & 13 \\
         Swedish   & KB-BERT \cite{malmsten2020playing}               & sv        & 50,325   & 18 \\
         Dutch     & BERTje \cite{de2019bertje}                & nl        & 30,000   & 12 \\
         Finnish   & FinBERT \cite{virtanen2019multilingual}               & fi        & 50,105   & $\approx$ 48 \\
         \multirow{2}{*}{Arabic}   & AraBERTv01 \cite{antoun2020arabert} & ar1  & 64,000   & 23 \\
                & AraBERTv02 \cite{antoun2020arabert} & ar2   & 64,000 & 77 \\
         \hline
         \end{tabular}
\caption{List of the monolingual BERT models considered. Data size refers to the size of data used for pre-training.}
\label{tab:models_list}
\end{table*}

\section{Method and Models}
\label{sec:method} {
Due to the lack of standardized monolingual downstream tasks in non-English languages, we have chosen to transfer from other languages into English. However, our method is likely to be most useful in a low-resource scenario. The proposed method is illustrated in Figure \ref{fig:method}. 
For each language pair, we take a pre-trained language model in a source language and \emph{ replace} the source language vocabulary with the English vocabulary and \emph{continue} pre-training the model on English data. We denote a model transferred with our method from the source language \textit{lang} to English by [\textit{lang}$\rightarrow$ en]. Given that the vocabulary tokens learned by a Wordpiece tokenizer are ordered by descending frequency, our method maps the vocabulary of the target language to the trained weights of the source embeddings with similar frequencies. In other words, the fifth most frequent token of the English vocabulary is initialised with the source embedding of the fifth most frequent token of the source vocabulary.

Our method is closely related to the MonoTrans method proposed by \newcite{artetxe-etal-2020-cross}. In both cases, cross-lingual transfer is primarily performed by adapting the embedding layer of a monolingual model. In \cite{artetxe-etal-2020-cross}, the embedding layer is learned from scratch in the target language, while the rest of the model parameters are frozen. In our case, we continue pre-training all model parameters in the target language, including the embedding layer. An additional difference is the method used for evaluating the model after transfer: we fine-tune the model on tasks in the target language, while \newcite{artetxe-etal-2020-cross} perform zero-shot evaluation.


\begin{figure}
    \centering
    \includegraphics[width=0.45\textwidth]{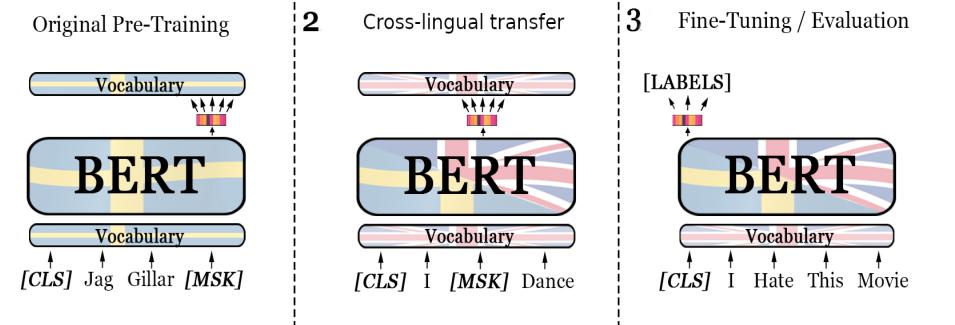}
    \caption{Our cross-lingual transfer method, applied to the transfer from Swedish to English: we continue pre-training Swedish BERT on our English Corpus, using the English vocabulary. Then, we fine-tune the transferred model on an English task.}
    \label{fig:method}
\end{figure}

The main criteria for including a monolingual language model in our experiments is that its architecture and training procedure should follow BERT-base \cite{devlin-etal-2019-BERT} and its pre-training corpus should include Wikipedia. The complete list of monolingual BERT models employed is presented in Table \ref{tab:models_list}, while the details of model selection can be found in Appendix section \ref{sec:model_selection}. 
Note that we have selected languages with varying degrees of linguistic similarity to English, and also one language with two different models with different amounts of training data.
}

\section{Experiments}

{\label{subsec:setup}
\subsection{Experimental Setup} 
Our pre-training corpus is English Wikipedia,\footnote{Downloaded in November 2019, using this script: \url{https://github.com/facebookresearch/XLM/blob/master/get-data-wiki.sh} } which amounts to $13$G English text. We train a Wordpiece tokenizer on the downloaded English Wikipedia, similar to \newcite{devlin-etal-2019-BERT}. The vocabulary size is fixed to $32$K. For [\textit{lang}] models with a larger vocabulary size, the vocabulary is resized to $32$K by keeping only the first $32$K vocabulary tokens. 

Each [\textit{lang}$\rightarrow$ en] model is trained for one epoch on English Wikipedia using the masked language modeling training objective. The complete list of training hyperparameters is shared between all trained models and can be found in Appendix section \ref{sec:training_details}.

All models are fine-tuned on the tasks included in the GLUE benchmark \cite{wang2018glue}, excluding WNLI given the known issues with the dataset construction.\footnote{\url{https://gluebenchmark.com/faq}(Number $12$)} The standard fine-tuning procedure is followed \cite{devlin-etal-2019-BERT} and the hyperparameters can be found in Appendix section \ref{sec:training_details}. We refrain from reporting the GLUE test results, since the comparison with models achieving SOTA performance in GLUE is not relevant for this work.

For each language pair (\textit{lang}, en), the fine-tuning performance of model [\textit{lang}$\rightarrow$ en] is compared with the performance of a BERT-base model, namely [en], that is trained from scratch on our English corpus using the exact same setup with the transferred models. This baseline allows us to compare the effect of choosing different initializations, namely random or trained monolingual models, when training a model in a new language. 

}

\subsection{GLUE Fine-Tuning}
\label{subsec:glue} {
\begin{table*}[htbp]
\centering
\begin{adjustbox}{width=\textwidth}
\begin{tabular}{|l|r|r|r|r|r|r|r|r|r|r|}
\hline
{\bf Lang} &  {\bf CoLA} &  {\bf MNLI} (m/mm) &  {\bf MRPC} &   {\bf QNLI} &   {\bf QQP} &  {\bf RTE} &  {\bf SST-2} & {\bf STS-B} & {\bf AVG} \\
 \hline
  en                       & 25.68 & 76.21/76.13 &  82.69 &  85.86 &   85.21 &  54.21 &  88.04 &  82.71 &        72.97   \\\hline \hline
  sv$\rightarrow$ en      &  43.65 & 80.72/81.77 &  88.93 &  89.11 &   86.32 &  55.08 &  90.22 &  84.91 &   {\bf 77.85}  \\\hline
  nl$\rightarrow$ en      & 39.87 & 78.96/79.79 &  85.65 &  87.34 &   85.82 &  55.01 &  89.10 &  83.65 &   {\bf 76.13}  \\\hline
  fi$\rightarrow$ en     & 40.01 & 79.90/80.52 &  87.82 &  88.30 &   86.37 &  52.12 &  88.18 &  83.82 &   {\bf 76.34}  \\\hline
  ar1$\rightarrow$ en    & 33.29 & 78.90/79.38 &  87.16 &  87.46 &   86.09 &  54.21 &  88.87 &  84.46 &   {\bf 75.54}  \\\hline
  ar2$\rightarrow$ en     & 39.82 & 79.52/80.28 &  88.46 &  88.35 &   85.72 &  57.18 &  90.10 &  83.77 &   {\bf 77.02} \\\hline
\end{tabular}
\end{adjustbox}
\caption{GLUE validation metric score for all models and tasks. Accuracy is the reported metric for all tasks, except from CoLA (Mathew correlation coefficient), QQP and MRPC (F1 score) and STS-B (Spearman correlation (x$100$)). The ``AVG'' column corresponds to the average score computed across all evaluated GLUE tasks. The bold font underlines the overall best performing model per source language, comparing to the [en] model.}
\label{tab:dev_glue_results}
\end{table*}

The GLUE validation results of our models are reported in Table \ref{tab:dev_glue_results}. The standard deviation of the metric score in each task is presented in Appendix section \ref{sec:apx_glue_results}. The results confirm that transferring a non-English model to English, namely [\textit{lang}$\rightarrow$ en], leads to significantly better fine-tuning performance than an English model trained from scratch, namely [en]. The best performing model, in terms of average score, is [sv$\rightarrow$ en], outperforming [en] by $4.88$ absolute difference. Additionally, we make the following observations:

\begin{itemize}

    \item{{\bf All transferred models improve over the [en] model independently of source language:}} Interestingly, the linguistic similarity between the source language and English does not significantly impact the effectiveness of our cross-lingual transfer method. 

    \item{{\bf The data size of the pre-training corpus in the source language matters:}} The English model transferred from [ar2], namely [ar2$\rightarrow$ en], performs better than [ar1$\rightarrow$ en], which originates from [ar1]. Given that [ar2] is trained on $\approx 3$ times more data than [ar1], this possibly indicates the important role of pre-training data size in cross-lingual performance. 

\end{itemize}
We also investigate the case where we apply English pre-training on the source monolingual models but \emph{keep} the source vocabulary. The fine-tuning results on GLUE, presented in Table \ref{tab:dev_glue_results_vocab}, demonstrate the importance of matching the tokenizer to the vocabulary of the target language. However, it is noteworthy that the models perform similarly (or even better) to the English model trained from scratch.

\begin{table}[!htbp]
\centering
\begin{adjustbox}{max width=0.35 \textwidth}
\begin{tabular}{|l|l|l|}
\hline
{\bf Lang} & {\bf Vocab} & {\bf AVG} \\
 \hline
  en                       & en  & 72.97   \\\hline \hline
  sv$\rightarrow$ en & sv     &  73.25          \\
  sv$\rightarrow$ en & en     &  77.85  \\\hline
  nl$\rightarrow$ en & nl    &        72.99  \\
  nl$\rightarrow$ en & en     &   76.13  \\\hline
  fi$\rightarrow$ en & fi      &        68.54   \\
  fi$\rightarrow$ en  & en    &   76.34 \\\hline
  ar1$\rightarrow$ en & ar1   &        73.99  \\
  ar1$\rightarrow$ en  & en  & 75.54 \\\hline
  ar2$\rightarrow$ en & ar2   & 74.27          \\
  ar2$\rightarrow$ en & en   &   77.02 \\\hline
\end{tabular}
\end{adjustbox}
\caption{Average GLUE validation score for all models, using the original or the English vocabulary. The bold font underlines the overall best performing model per source language.}
\label{tab:dev_glue_results_vocab}
\end{table}
}

\subsection{English Linguistic Probing}
\label{subsec:probing} {
We also study the linguistic effects of the proposed cross-lingual transfer method. This is done by evaluating the syntactic and semantic knowledge of the [en] compared to the [lang$\rightarrow$ en] models through probing their representations. 

More specifically, we evaluate the word representations yielded by [lang$\rightarrow$ en] using the \emph{structural probe} model, proposed by \newcite{hewitt-manning-2019-structural}, which detects whether syntactic trees are encoded in a linear transformation of the model embedding space. In this way, we evaluate if the word embeddings of the transferred English models encode syntactic parsing information. Following \newcite{hewitt-manning-2019-structural}, we define the structural probe model as a linear transformation that learns the tree distances between all pairs of words in training sentences from the English part of Universal Dependencies v2.7 (English-EWT).\footnote{\url{https://github.com/UniversalDependencies/UD_English-EWT}} The trained probing model is then evaluated on the English-EWT test set using the following evaluation metrics \cite{hewitt-manning-2019-structural}: Spearman correlation between predicted and true word pair distances (DSpr), averaged across the input sentences with length $5$-$50$, and the percentage of undirected edges placed correctly in comparison with the gold parse tree, namely undirected unlabelled attachments score (UUAS).

\begin{table*}[htbp]
\centering
\begin{tabular}{|l|r|r||r|}
\hline
  &\multicolumn{2}{c||}{\textit{Syntax}} & \textit{Semantics}\\
  \cline{2-4}
\textbf{Language} & \textbf{UUAS}& \textbf{DSpr} & \textbf{WiC} (acc)  \\ \hline
en                       & 66.21 & 70.41 & 56.73 \\\hline \hline
sv$\rightarrow$ en      & 67.22 & 72.15 & 61.09 \\\hline
nl$\rightarrow$ en      &  66.59 & 71.73 & 59.46 \\\hline
fi$\rightarrow$ en     & 67.02 & 71.56 & 61.06 \\\hline
ar1$\rightarrow$ en    & 67.53 &71.67 & 59.99 \\\hline
ar2$\rightarrow$ en     & 64.98 &  70.48 & 59.90 \\\hline
\end{tabular}

\caption{Probing results of the [en] and English transferred models on the English-EWT test set using the structural probe model \protect\cite{hewitt-manning-2019-structural} and on the WiC \protect\cite{pilehvar-camacho-collados-2019-wic} dev set.}
\label{tab:english_probing}
\end{table*}

For the semantic probing of the transferred models, we use the Words In Context task (WiC \newcite{pilehvar-camacho-collados-2019-wic}). This is a binary classification task, where the model needs to determine if a given word is used with the same meaning or not in two different contexts. For this purpose, we train a linear classifier on top of each sentence representation. The details of our probing setup can be found in Appendix section \ref{sec:probing_setup}.

The probing results are presented in Table \ref{tab:english_probing}. The improvement of the [\textit{lang}$\rightarrow$ en] models over the [en] model on the WiC test for all source languages demonstrates that semantic abstractions learned in the source language are transferred to English and enhance probing performance. Interestingly, the results of syntactic probing on the transferred models are in the best case similar to the probing results on the [en] model. It is worth observing that unlike the results on the GLUE benchmark, [ar2$\rightarrow$ en] performs clearly worse than [ar1$\rightarrow$ en] on syntactic probing. This indicates that larger pre-training data size hinders the learning of syntactic information in the target language through language pre-training. Overall, the results on English probing indicate that our cross-lingual transfer method boosts the learning of semantic information in the target language, but does not enhance the learning of syntactic information.


}

\section{Discussion}
The monolingual experimental setup employed here allows us to control for the source language and study the effect of this choice on cross-lingual performance. The presented results on GLUE as well as English probing tasks show that the linguistic similarity between source and target language is not important for cross-lingual transfer in our setup. This result contradicts previous work by \newcite{lauscher-etal-2020-zero} that studies the effect of language similarity in zero-shot cross-lingual transfer and finds that multilingual language models have poor zero-shot performance in distant target languages. 


Our setup differs significantly from \emph{zero-shot} cross-lingual transfer, where the source model is transferred to the target language at test time, after being fine-tuned on a task in the source language. By contrast, we adapt trained monolingual models to the target language through additional pre-training. This is inspired by a domain adaptation approach proposed by \newcite{gururangan-etal-2020-dont}. They show that adapting the original language model to the target domain through extra pre-training improves model performance on the target task. Our results suggest that this approach is also beneficial for model transfer between two different language spaces. Future work will study the adaptation of multilingual models with our method and perform a comparison with zero-shot cross-lingual transfer.

As part of the proposed cross-lingual transfer method, each token embedding in the target embedding matrix is initialised with the trained source embedding at the same position in the source embedding matrix. The investigation of the effect of this frequency-based embedding initialisation scheme on model performance in the target language, in comparison with other types of initialisations such as random, is left for future work.

Due to the lack of established evaluation benchmarks in other languages, we perform cross-lingual transfer experiments only from other languages to English. However, we believe that our method can provide a smart initialization for training models in minority languages, where neither large amounts of data nor computational resources are available. In this direction, \newcite{de-vries-etal-2021-adapting} transfer monolingual BERT models to two Dutch dialects using zero-shot learning. 

\section{Conclusion}
In this paper, we show that using a pre-trained language model as initialization for pre-training in new language spaces is beneficial with regard to model performance on downstream tasks and linguistic knowledge in the target language. Our experimental results demonstrate that language similarity has no impact on cross-lingual performance, while larger pre-training data size appears to have a positive effect on monolingual task performance in the target language.

We hope that our work will inspire practitioners and researchers to initialize new monolingual models with parameters from existing models if possible, in order to reach competitive models in low resource settings and reach better performance on downstream tasks.

\section{Acknowledgements}
This work is supported by the Swedish innovation agency (Vinnova) under contract 2019-02996. We would like to thank Joakim Nivre for his useful feedback on this work and Fredrik Carlsson for the illustration of the method (Figure 1).

\section{References}
\bibliographystyle{lrec2022-bib}
\bibliography{main}

\clearpage

\appendix

\begin{table*}[htbp]
\centering
\begin{adjustbox}{width=1.00\textwidth}
\begin{tabular}{|l|r|r|r|r|r|r|r|r|r|}
\hline
{\bf Lang} & {\bf CoLA (stdev)} &  {\bf MNLI (stdev)} (m/mm) &  {\bf MRPC (stdev)} &   {\bf QNLI (stdev)} &   {\bf QQP (stdev)} &  {\bf RTE (stdev)} &  {\bf SST-2 (stdev)} & {\bf STS-B (stdev)}   \\
 \hline
  en                       &  2.62 &   0.23/0.28 &  0.42 &  0.30 &  0.05 &  1.89 &    0.85 &  0.19  \\\hline \hline
  sv$\rightarrow$ en     & 1.21 &   0.17/0.09 &  0.19 &  0.19 &  0.07 &  1.95 &    0.32 &  0.28  \\\hline
  nl$\rightarrow$ en     & 1.22 &   0.24/0.27 &  0.75 &  0.44 &  0.09 &  0.82 &    0.36 &  0.40 \\\hline
  fi$\rightarrow$ en     & 2.71 &   0.12/0.14 &  0.65 &  0.28 &  0.09 &  1.54 &    0.51 &  0.48 \\\hline
  ar1$\rightarrow$ en    &  2.63 &   0.16/0.18 &  0.71 &  0.54 &  0.10 &  2.87 &    0.68 &  0.15 \\\hline
  ar2$\rightarrow$ en     & 0.74 &   0.19/0.21 &  1.06 &  0.37 &  0.05 &  2.12 &    0.43 &  0.44 \\\hline
  
\end{tabular}
\end{adjustbox}
\caption{Standard deviation in GLUE validation results. Each value corresponds to the square root of the sample variance from the mean, computed across $5$ runs per combination of model and task.}
\label{tab:glue_deviation}
\end{table*}

\section{Selection of monolingual language models}
\label{sec:model_selection}{
All monolingual models used were downloaded from the HuggingFace model hub,\footnote{\url{https://huggingface.co/}} which is an open library. All selected models are cased, with the exception of AraBERTs. The mapping between alias name in the paper and model name in the Hugging Face model hub is presented in Table \ref{tab:models_hf}.

In Table \ref{tab:models_list}, the pre-training data sizes of KB-BERT, BERTje and AraBERTs were taken directly from the corresponding papers \cite{malmsten2020playing,de2019bertje,antoun2020arabert}. For BERT-base, our estimation of the pre-training data size is based on the total number of words ($3.3$B) in the pre-training corpus, provided by \cite{devlin-etal-2019-BERT}. Our estimation for FinnBERT is based on the total number of characters ($24$B), stated in the original paper \newcite{virtanen2019multilingual}.

\begin{table}[htbp]
 \centering
 \begin{adjustbox}{max width=0.45\textwidth}
        \begin{tabular}{|l|l|}
        \hline
         \textbf{Alias} & \textbf{Hugging Face name} \\
         \hline
         BERT-base (original) & bert-base-cased \\
         sv        & KB/bert-base-swedish-cased \\
         nl        & GroNLP/bert-base-dutch-cased \\
         fi        & TurkuNLP/bert-base-finnish-cased-v1 \\
         ar1  & aubmindlab/bert-base-arabertv01 \\
         ar2   & aubmindlab/bert-base-arabertv02 \\
         \hline
         \end{tabular}
\end{adjustbox}
\caption{Mapping between model alias name in the paper and model name in the Hugging Face model hub.}
\label{tab:models_hf}
\end{table}

}

\section{Training details}
\label{sec:training_details} {
The implementation of English pre-training and GLUE fine-tuning is heavily based on the example scripts\footnote{\url{https://github.com/huggingface/transformers/blob/master/examples/pytorch/language-modeling/run_mlm.py}, \url{https://github.com/huggingface/transformers/blob/master/examples/pytorch/text-classification/run_glue.py}} provided by the HuggingFace \texttt{transformers} library. The hyperparameters used are presented in Table \ref{tab:hyperparams}. The parameters which are not presented here have been set to their default value.\footnote{\url{https://huggingface.co/transformers/_modules/transformers/training_args.html}} For training the English Wordpiece tokenizer, we used the example code which is part of the HuggingFace \texttt{tokenizers} library.\footnote{\url{https://huggingface.co/docs/tokenizers/python/latest/pipeline.html\#all-together-a-bert-tokenizer-from-scratch}} All models were trained on a single GPU machine (Nvidia Tesla v100 sxm2 32GB), with the average training time per model being $4$ days.

\begin{table}[htbp]
\centering
\begin{adjustbox}{max width=0.45\textwidth}
\begin{tabular}{|l|r|r|}
\hline
\textbf{Hyperparameter} & \textbf{Training value} & \textbf{Fine-tuning value} \\
\hline
batch size & 128 & 32\\
learning rate & 5e-5 & 2e-5\\
maximum sequence length & 128 & 128 \\ 
train\_epochs  & 1.0 & 3.0 \\
optimizer & AdamW & AdamW \\
\hline
\end{tabular}
\end{adjustbox}
\caption{Hyperparameters used for English pre-training (second column) and GLUE fine-tuning (third column).}
\label{tab:hyperparams}
\end{table}

}

\section{GLUE results}
\label{sec:apx_glue_results} { 

For each model and task, we perform $5$ runs with varying random seeds. The standard deviation from the average performance across these $5$ runs is presented in Table \ref{tab:glue_deviation}. Overall, we observe that the standard deviation is notably higher in CoLA and RTE comparing to the rest of the GLUE tasks.

}

\section{English linguistic probing experiments}
\label{sec:probing_setup} {
\subsection{Syntactic probing}
Our implementation of the structural probing method \cite{hewitt-manning-2019-structural} is heavily based on the coding repositories\footnote{\url{https://github.com/ethanachi/multilingual-probing-visualization} \url{https://github.com/john-hewitt/structural-probes/}}  of \cite{hewitt-manning-2019-structural,chi-etal-2020-finding}. 

\subsection{Semantic probing}
\label{subsec:semantic_prob_setup}{

The \texttt{jiant} library\footnote{\label{jiant}\url{https://github.com/nyu-mll/jiant}} was used for training and evaluation of our models on the WiC task \cite{pilehvar-camacho-collados-2019-wic}. The reported accuracy corresponds to the average accuracy across $5$ runs, each one with a different random seed. The standard deviation across runs is presented in Table \ref{tab:wic_deviation}.

\begin{table}[htbp]
\centering

\begin{adjustbox}{max width=0.45\textwidth}
\begin{tabular}{|l|c|}
\hline
\textbf{Lang}  & \textbf{WiC (stdev)} \\ \hline
en                    & 1.19  \\\hline \hline
sv$\rightarrow$ en   & 1.27 \\\hline

nl$\rightarrow$ en    & 0.76 \\\hline

fi$\rightarrow$ en  & 1.76 \\\hline

ar1$\rightarrow$ en   & 1.10 \\\hline

ar2$\rightarrow$ en  & 1.22 \\\hline

\hline
\end{tabular}
\label{tab:wic_deviation}
\end{adjustbox}
\caption{Standard deviation (stdev) of the average accuracy on WiC validation set over 5 runs with different random seeds. Standard deviation is computed as the square root of the sample variance from the mean accuracy.}
\end{table}
}

}

\end{document}